\RequirePackage{fix-cm}
\documentclass[conference, final]{IEEEtran}
\usepackage[T1]{fontenc}
\usepackage[utf8]{inputenc}
\usepackage{graphicx}
\usepackage{amsmath,amssymb}
\usepackage{hyperref}
\usepackage{tikz}
\usepackage{environ}
\usetikzlibrary{shapes,arrows,positioning,calc}

\begin{document}

\title{On Policy Gradients}

\author{
  \IEEEauthorblockN{Mattis Manfred K{\"a}mmerer} 
  \IEEEauthorblockA{
    \textit{Technische Universit{\"a}t Darmstadt} \\
    \href{https://orcid.org/0000-0002-6869-2379}{orcid.org/0000-0002-6869-2379}
  }
}

\maketitle

\begin{abstract}

The goal of policy gradient approaches is to find a policy in a given class of policies which maximizes the expected return.
Given a differentiable model of the policy, we want to apply a gradient-ascent technique to reach a local optimum. 
We mainly use gradient ascent, because it is theoretically well researched.
The main issue is that the policy gradient with respect to the expected return is not available, thus we need to estimate it.
As policy gradient algorithms also tend to require on-policy data for the gradient estimate, their biggest weakness is sample efficiency.
For this reason, most research is focused on finding algorithms with improved sample efficiency.
This paper provides a formal introduction to policy gradient that shows the development of policy gradient approaches, and should enable the reader to follow current research on the topic.

\end{abstract}

\section{Introduction}
\label{intro}

Policy gradient methods are approaches to maximize the expected return in a Markov Decision Process (MDP). 
Using a parameterized policy to decide the next action, they can easily incorporate prior domain knowledge, but require a lot of configuration to produce an effective agent for a specific environment. 
Also, they frequently require on-policy training and a lot of samples to find a good policy. 
In this paper, we focus on approaches to estimate the policy gradient, though we introduce the most common policy classes shortly.
Policy improvement means a step in the parameter space such that the policy under the new parameters will on average perform better than the old policy, i.e. improve its expected return. 
Policy gradient estimation is the term we use to describe the process of computing the direction in parameter space for a policy improvement. 
Essentially, the goal is to estimate the gradient of the policy with respect to the expected return. 
Since this is the core problem of policy gradient methods, it is also the main topic of this paper.

In section \ref{sec:prel}, we give some preliminaries and describe the problem setup in detail. 
In section \ref{sec:pge}, we discuss different approaches to estimate the policy gradient. 
Using our insights from section \ref{sec:pge}, we derive the actor-critic framework in section \ref{sec:ac}, which harnesses value-function estimation for improved gradient updates. 
Then, in section \ref{sec:natural}, we introduce some gradient-ascent methods that build on the approaches given in sections \ref{sec:pge} and \ref{sec:ac}, refining the gradient estimation by Fisher's information matrix to get the natural gradient. 
Finally, in section \ref{sec:outro}, we summarize the contents presented in this paper, and give a short conclusion.

\section{Preliminaries}
\label{sec:prel}

We define states $s \in \mathbb{S}$, actions $a \in \mathbb{A}$, and rewards $r \in \mathbb{R}$. 
A trajectory $\tau := (s_0, a_0, $ $s_1, a_1, \dots, s_T, a_T)$ is generated by drawing $s_0 \sim \mu_0(s_0)$ according to the distribution over initial states $\mu_0(s_0)$, and successively sampling $a_t \sim \pi(a_t|s_t)$ according to the policy $\pi$ parameterized by $\theta$, and $s_{t+1} \sim p(s_{t+1}|s_t,a_t)$ until the horizon $T$, or a terminal state is reached. 
At each time step, we receive a reward according to $r_t = r(s_t, a_t) \equiv \mathbb{E}_{s'}\left[r(s_t,a_t,s')\right]$. 
A trajectory can also be called roll-out or episode, though the term episode implies it ends in a terminal state.
We assume a Markov Decision Process (MDP), meaning the probability distribution of the next states is independent of past states $s_{0:t-1}$ given the present state $s_t$ and action $a_t$, 
\begin{equation}
	p(s_{t+1}|s_t,a_t)=p(s_{t+1}|s_{0:t},a_{0:t}).
\end{equation}
Where we define $i:j$ with $i,j \in \mathbb{N}, i < j$ as an index over all integers from $i$ to $j$, i.e., $s_{i:j} \equiv s_i, s_{i+1}, \dots, s_j$. 
We assume no additional prior knowledge about the environment, meaning we assume the probability of a trajectory is 
\begin{equation}
	p_\pi(\tau) = \mu_0(s_0) \prod_{t=0}^{T-1} p(s_{t+1}|s_t, a_t) \pi(a_t|s_t).
\end{equation}

The most frequently used policy classes in policy gradient approaches are Gibbs policies $\pi(a|s) = \frac{\exp(\phi(s,a)^T\theta)}{\sum_b \exp(\phi(s,b)^T\theta)}$ \cite{Sutton:1999:PGM:3009657.3009806,Bagnell2004LearningD} for discrete problems, and Gaussian policies $\pi(a|s) = \mathcal{N}(\phi(s,a)^T\theta_1,\theta_2)$ for continuous problems, where $\theta_2$ is an exploration parameter \cite{Williams92simplestatistical,peter:article:1996}, and $\phi(s,a)$ is the vector of basis functions on the state-action pair.

\paragraph{Policy gradient} 
Our goal with respect to episodes is to maximize the expectation of the total reward, also called expected return. 
The total reward in the horizon $T$ is $\sum_{t=0}^{T} r_{t}$. 
We additionally introduce a discount factor $\gamma \in [0,1)$.
Intuitively, this reflects the idea that the relevance of later actions declines, and ensures that the return is finite, even for the infinite horizon $T \to \infty$. 
The discounted total reward is 
\begin{equation}
  \mathcal{R}^\tau \equiv \mathcal{R}_0^T := \sum_{t=0}^{T} \gamma^t r_t.
  \label{eqn:acc-reward}
\end{equation}

Since we have only limited knowledge of the performance of the policy, we need to approximate an optimal policy by estimating a gradient. 
Thus, we search $\nabla_\theta J(\theta) := \nabla_\theta \mathbb{E}_{p_\pi(\tau)}\left[\mathcal{R}^\tau\right]$, to make a policy gradient step according to $\theta_{k+1} = \theta_k + \alpha_k \nabla_\theta J(\theta)$, where $\alpha_k$ denotes a learning rate. 
Section \ref{sec:pge} shows how we can estimate $J(\theta)$.

\section{Policy Gradient Estimation}
\label{sec:pge}

In this section, we introduce methods for estimating the policy gradient.

\paragraph{Finite-difference gradients} 
A simple approach for gradient estimation is to choose a small $\delta\theta$, and evaluate the new policy given the slightly changed parameters as in 
\begin{equation}
	\nabla_\theta J(\theta) \approx \frac{J(\theta+\delta\theta)J(\theta-\delta\theta)}{2\delta\theta}.
\end{equation} 
This can lend a good estimate of the gradient given a small $\delta \theta$, and is generally called the symmetric derivative. 
However, finite-difference gradients suffer from the curse of dimensionality, and can require very small $\delta\theta$.
Thus, finite-difference gradients only work well in specific scenarios, but should not be discarded due to simplicity.

\paragraph{Value functions} 
Given we know the actual value of a state, i.e. the expected return we will get starting from state $s_t$, this function can be used to evaluate the performance of our policy, and can be written as
\begin{equation}
	V^{\pi}(s_t) := \mathbb{E}_{\substack{s_{t+1:h} \\ a_{t:h}}}\left[\mathcal{R}_t^T\right].
	\label{eqn:v}
\end{equation}
In addition to the value function, we also define the state-action value function, often called Q-function. 
Instead of the expected accumulated reward starting from state $s_t$, this function gives the expected accumulated reward given an action $a_t$ is selected in state $s_t$, 
\begin{equation}
	Q^{\pi}(s_t, a_t) := \mathbb{E}_{\substack{s_{t+1:h} \\ a_{t+1:h}}}\left[\mathcal{R}_t^T\right].
	\label{eqn:q}
\end{equation}
As we will see, this function also gives us the true gradient of $J(\theta)$, though in general we need to estimate it. 
Using the value function, and the Q-function, we can derive a better estimate of the policy gradient.

\paragraph{Likelihood-ratio gradients} 
For this derivation, we will change the perspective a bit, requiring some additional definitions. 
We define ${\mu_\pi}_i = \sum_{t=0}^{\infty} \gamma^t p(s_t=s_i | s_0, \pi)$ as the discounted state distribution, though it does not sum up to one without normalization, which can be achieved by multiplying by $(1-\gamma)$. 
Note that $\mu_\pi$ is equivalent to the discounted state visit count $d^\pi$ introduced by Sutton et al. \cite{Sutton:1999:PGM:3009657.3009806}.
Further, we define $P_\pi$ as the transition matrix, i.e. ${P_\pi}_{i,j}=\sum\nolimits_k p(s_j|s_i,a_k)\pi(a_k|s_i)$, $r_\pi$ as the mean rewards for all states given by ${r_\pi}_i = \sum_j r(s_i,a_j)\pi(a_j|s_i)$, and $\mu_0 = \left[\mu_0(s_0),\mu_0(s_1),\ldots\right]^T$ as a vector representing the initial state distribution.
Finally, we define $V_\pi = \left[ V_\pi(s_0), V_\pi(s_1), \ldots \right]^T$, from which it follows that $V_\pi = \mu_\pi r_\pi$, so we can reformulate the problem as 
\begin{align}
	\underset{\theta}{\text{max}}\ J(\theta) &= \mu_0^T V_\pi \\
	\text{s.t.}\ V_\pi &= r_\pi + \gamma P_\pi V_\pi . \nonumber
\end{align}
Since $\mu_0^T$ does not depend on $\theta$, 
\begin{align*}
	\nabla_\theta J(\theta) &= \nabla_\theta \mu_0^T V_\pi = \mu_0^T \nabla_\theta V_\pi.
\end{align*}
We can replace $\nabla_\theta V_\pi$ using
\begin{align*}
	\nabla_\theta V_\pi &= \nabla_\theta \left( r_\pi + \gamma P_\pi V_\pi \right)
	\\ \nabla_\theta V_\pi &= \nabla_\theta r_\pi + \gamma (\nabla_\theta P_\pi) V_\pi + \gamma P_\pi \nabla_\theta V_\pi
	\\ (I-\gamma P_\pi) \nabla_\theta V_\pi &= \nabla_\theta r_\pi + \gamma (\nabla_\theta P_\pi) V_\pi
	\\ \nabla_\theta V_\pi &= (I - \gamma P_\pi)^{-1} (\nabla_\theta r_\pi + \gamma (\nabla_\theta P_\pi) V_\pi), 
\end{align*}
and find that
\begin{align*}
	\mu_\pi &= \mu_0 + \gamma P_\pi^T \mu_\pi \nonumber \\
	(I- \gamma P_\pi^T)\mu_\pi &= \mu_0 \nonumber \\
	\mu_\pi^T &= \mu_0^T (I- \gamma P_\pi)^{-1} ,
\end{align*}
which we can take back into the gradient equation 
\begin{align}
	\nabla_\theta J(\theta) &= \mu_0^T (I - \gamma P_\pi)^{-1} (\nabla_\theta r_\pi + \gamma (\nabla_\theta P_\pi) V_\pi) \nonumber
	\\ &= \mu_\pi^T (\nabla_\theta r_\pi + \gamma (\nabla_\theta P_\pi) V_\pi) \nonumber
	\\ &\equiv \sum\nolimits_{i,j} \mu(s_i) \nabla_\theta\pi(a_j|s_i) Q_\pi(s_i,a_j) \label{eqn:likelihood-equiv}
	\\ &= \sum\nolimits_{i,j} \mu(s_i) \pi(a_j|s_i) \nabla_\theta\log\pi(a_j|s_i) Q_\pi(s_i,a_j). \nonumber
\end{align}
The equivalence in \eqref{eqn:likelihood-equiv} comes from the observation that $\nabla_\theta \pi(a|s)$ is distributed over the addition $\nabla_\theta r_\pi + \gamma (\nabla_\theta P_\pi) V_\pi$.
When we take out this common factor, $Q_\pi(s,a)$ remains.
Then, we use $\nabla_\theta \pi(a|s) = \pi(a|s)\nabla_\theta\log\pi(a|s)$, obtained from the likelihood ratio $\nabla_\theta \log p(x|\theta) = \frac{\nabla_\theta p(x|\theta)}{p(x|\theta)}$. 
This gives us the likelihood-ratio gradient
\begin{equation}
	\nabla_\theta J(\theta) = \mathbb{E}_{\substack{\ s \sim \mu_\pi \\a \sim \pi}} \Big[\nabla_\theta{\log\pi(a|s)}Q_\pi(s,a)\Big],
	\label{eqn:like-grad}
\end{equation}
intuitively meaning we should increase the probability of actions that lead to higher Q-values.
This formulation enables us to calculate the gradient $\nabla_\theta J(\theta)$, while directly taking advantage of the MDP structure in the form of $Q_\pi(s,a)$.

Obviously, we do not have the true $Q_\pi(s,a)$, thus we need to approximate it by $\hat{Q}_\pi(s,a)$. 
In all of the following sections, when we say that we sample an episode, we mean to draw $a \sim \pi(a|s)$, starting in state $s_0 \sim p(s_0)$ and match a function estimator to our observations.
Following this procedure, it is shown that for $\lim_{k\to\infty}\alpha_k = 0$, and $\sum_{k=0}^\infty \alpha_k$ we are guaranteed to converge to a local optimum \cite{Sutton:1999:PGM:3009657.3009806}.
Approximating $Q_\pi(s,a)$ by an unbiased estimator $f_w^\pi(s_t, a_t) \equiv \hat{Q}_\pi(s_t, a_t)$, Sutton et al. \cite{Sutton:1999:PGM:3009657.3009806} show that using this function approximation we will converge to the true local optimum of $J(\theta)$.

\paragraph{Episode-based updates} 
A very general optimization approach to this optimization problem are episodic algorithms. 
We take a search distribution $p(\theta|\omega)$ over the parameter space of the policy class $\pi$, and sample acting policies from that distribution. 
The policy class $\pi$ is most often chosen deterministic. 
Using these policies, we sample trajectories $\tau$, and update the search policy using the returns of our sampled roll-outs
\begin{equation}
	\nabla_\theta J(\theta) \approx \sum_{t=0}^T \nabla_\omega \log p(\theta|\omega) \mathcal{R}_t^T.
\end{equation}
The resulting algorithms are black-box optimizers, and as such are largely applicable, but can not use any temporal information and have a lot of variance.
Given these insights, we require a way to design algorithms that improve the acting policy stepwise by observing each interaction with the environment.

\paragraph{Step-based updates}
The first class of algorithms developed to update a policy directly using a critic are called REINFORCE \cite{Williams92simplestatistical}.
REINFORCE samples a complete episode, at which point we can calculate the actual state-action value by traversing backwards over the trajectory, and estimates
\begin{equation}
	\nabla_\theta J(\theta) \approx \nabla_\theta\log\pi(s) \left(Q(s_t,a_t) - b_\tau \right).
	\label{eqn:reinforce}
\end{equation}
This is sometimes also called Monte-Carlo gradient estimation. 
However, given $b_\tau = 0, r_t > 0, \forall t=0,\dots,h$, we can only increase action probabilities. 
Obviously, we normalize to ensure $\forall s \in \mathbb{S}: \int_\mathbb{A}{\pi(a|s)da} = 1$. 
This means actions can only become less probable in relation to other actions. 
We find that this introduces more variance when learning from samples \cite{Sutton:1999:PGM:3009657.3009806}, and by that defeats the purpose of why we thought of this approach in the first place. 
One way to counter the variance is to use an effective baseline $b_\tau$.
Peters et al. \cite{4867} find that an estimate of the optimal baseline can be calculated by
\begin{equation}
	b_\tau = \frac
		{\left\langle 
			\left(\sum_{t=0}^T \nabla_\theta \log\pi(a_t|s_t) \right)^2 \sum_{t'=0}^T a_{t'} r_{t'} 
		\right\rangle}
		{\left\langle
			\left(\sum_{t=0}^T \nabla_\theta \log\pi(a_t|s_t) \right)^2
		\right\rangle},
\end{equation}
which does not affect the unbiasedness of the estimate.

Whenever we require estimating a value function for updating our policy, we can name the policy actor, and the estimated value function critic. 
From this observation, we define a class of policy optimization methods called actor-critic methods in section \ref{sec:ac}.

\section{Actor-Critic Methods} 
\label{sec:ac}

Policy gradient methods can be described in terms of two main steps often called policy evaluation and policy improvement. 
For actor-critic approaches, we separate these steps from the actor component by implementing a critic. 
This means, the actor consists only of the policy, while the critic is focused on estimating a score for the actions taken. 
By that concept, observations of the environment are given to the actor only to decide the next action, and to the critic only to improve its function estimation with the respective rewards. 
Figure \ref{fig:ac} shows the general structure of an actor-critic algorithm.
Given this definition, we can already say that the algorithms presented at the end of section \ref{sec:pge} are actor-critic approaches.

\tikzset{block/.style= {draw, rectangle, align=center, minimum height=2em, minimum width=3cm}}
\begin{figure}
  \begin{center}
  \begin{tikzpicture}[auto, node distance=2cm,>=latex']
    \node [block, name=env, minimum width=6cm] (env) {Environment};
    \node [block, above of=env, node distance=1.5cm] (critic) {Critic \\ $\hat{Q}_\pi(s,a)$};
    \node [block, above of=critic] (actor) {Actor \\ policy $\pi(a|s)$};
    
    \draw[densely dotted] ($(actor.north west)+(-1.5,0.6)$) rectangle ($(critic.south east)+(1.5,-0.15)$);
    \node[above of=actor, node distance=0.7cm] {Agent};

    \draw [->, align=left, swap, densely dashed] (critic.north) -- node{Policy \\ Improvement} (actor.south);
    \draw [->, pos=0.75] ($(env.north west)!0.15!(env.north)$) |- node{$s_t$} (actor.west);
    \draw [->, pos=0.75] ($(env.north west)!0.15!(env.north)$) |- node{$s_t, r_t$} (critic.west);
    \draw [->, pos=0.25] (actor.east) -| node{$a_t$} ($(env.north east)!0.15!(env.north)$);
    \draw [->, pos=0.75, swap] ($(env.north east)!0.15!(env.north)$) |- node{$a_t$} (critic.east);
    
    \draw [-, swap, pos=0.14] ($(env.north west)!0.15!(env.north)$) |- node{Observation} (critic.west);
    \draw [-, swap, pos=0.95] (actor.east) -| node{Action} ($(env.north east)!0.15!(env.north)$);
  \end{tikzpicture}
  \end{center}
  \caption{A visualization inspired by Kimura et al. \cite{Kimura1998AnAO}, showing the actor-critic framework.} \label{fig:ac}
\end{figure}
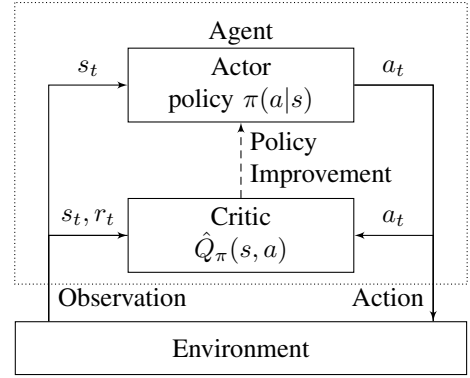

The critic estimates a state-action value function as defined in \eqref{eqn:q}. 
Sutton et al. \cite{Sutton:1999:PGM:3009657.3009806}, and Konda et al. \cite{Konda:2003:AA:942271.942292} find that the estimation $f_w^\pi(s,a) \approx Q_\pi(s,a)$ does not affect the unbiasedness of the gradient estimate under some restrictions. 
Specifically, this holds for 
\begin{equation}
	f_w^\pi(s,a) = {\nabla_\theta \log\pi(a|s)}^T w,
	\label{eqn:compatible-approximator}
\end{equation}
thus $f_w^\pi(s,a)$ being a linear function parameterized by the vector $w$.
Sutton et al. \cite{Sutton:1999:PGM:3009657.3009806} call this a compatible function approximator.
This guarantees that the function estimator does not cause divergence, and really enables recent research in reinforcement learning for continuous control problems, e.g., in humanoid robotics.

Traditionally, the improvement is often done by Monte-Carlo sampling as in REINFORCE \eqref{eqn:reinforce}, or using temporal difference (TD) \cite{Sutton1988}, i.e., we use the temporal difference between the critic's estimations 
\begin{equation}
  \delta(s_t) = r_t + \gamma \hat{V}_\pi(s_{t+1}) - \hat{V}_\pi(s_t).
  \label{eqn:td-error}
\end{equation}

However, Sutton et al. \cite{1993b} find that this is only guaranteed to be unbiased, if $\int_\mathbb{A}{\pi(s,a)f_w^\pi(s,a)da} = 0, \forall s \in \mathbb{S}$. 
Given this assumption, the function estimator $f_w^\pi$ is limited to approximating an advantage function
\begin{equation}
	f_w^\pi(s_t, a_t) \equiv \hat{A}_{\pi}(s_t, a_t) = \hat{Q}_{\pi}(s_t, a_t) - \hat{V}_{\pi}(s_t),
	\label{eqn:adv}
\end{equation} 
which requires bootstrapping for $\hat{V}_\pi$. 
If we use temporal difference in this context, we run into a problem, as \eqref{eqn:adv} subtracts $\hat{V}_\pi(s_t)$, meaning we would only learn immediate rewards \cite{Peters_IICHR_2003}. 
This would render the process biased. 
Sutton et al. \cite{Sutton:1999:PGM:3009657.3009806} and Konda et al. \cite{NIPS1999_1786} suggest estimating an action value function as in \eqref{eqn:q}. 
We can approximate this $f_w^\pi$ by least-squares optimization over multiple $\hat{Q}_\pi(s,a)$ obtained from roll-outs. 
However, Peters et al. \cite{4863} find that this approximation is highly reliant on the distribution of the training data. 
This comes from the realization, that we use only a subspace of the true action-value function in $\hat{V}^\pi$, which is only a state value function. 
One can compare this to approximating a parabola by a line, whereby the approximation changes wildly depending on which part of the parabola is in the training data. 
An approach to solve this bootstrapping problem is to rewrite the Bellman Equation using \eqref{eqn:adv} and \eqref{eqn:q}.
With $\hat{A}_\pi(s,a) = f_w^\pi(s,a)$, $\hat{V}_\pi(s) = \phi(s)^T v$, a zero-mean error term $\epsilon \equiv \epsilon(s_t,a_t,s_{t+1})$, we get
\begin{align}
  &\hat{A}_\pi(s,a) + \hat{V}_\pi(s) = r(s,a) + \gamma \int_\mathbb{S} p(s'|s,a)\hat{V}_\pi(s')ds', \\
  &\nabla_\theta \log \pi(a_t|s_t)^T w + \phi(s_t)^T v = r(s_t,a_t) + \gamma \phi(s_{t+1})^T v + \epsilon ,
\end{align}
which involves only linear equations to solve \cite{4863}.

With these insights in mind, section \ref{sec:natural} presents the natural gradient, a refined type of gradient which has a convenient fit in the actor-critic setting we just established.

\section{Natural Gradient}
\label{sec:natural}

Natural gradients were at first proposed for use in supervised learning settings by Amari et al. \cite{Amari:1998:NGW:287476.287477}, but have been shown to be effective in reinforcement learning by Kakade \cite{Kakade:2001} and Peters et al. \cite{4863}.

\begin{figure}
  \includegraphics[width=0.485\textwidth]{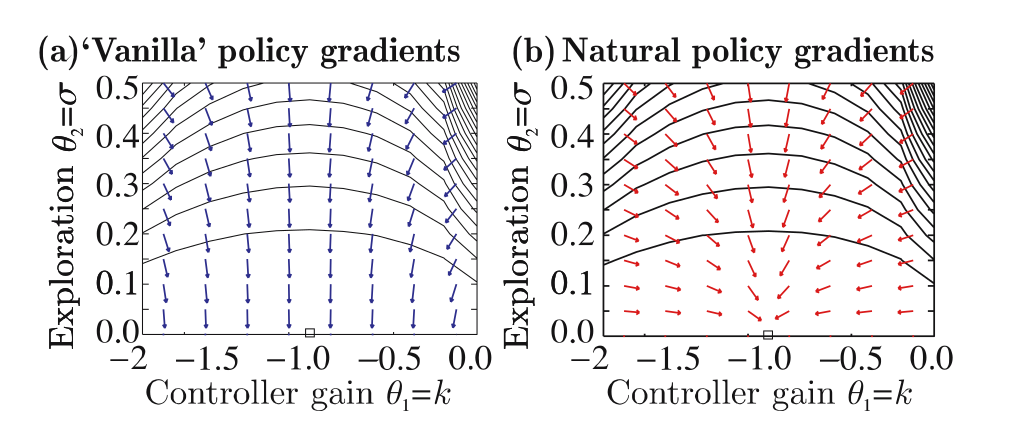}
  \caption{An experiment showing where the natural gradient has a great advantage \cite{Peters_IICHR_2003}. }\label{fig:nat-grad-adv}
\end{figure}

When using normal gradient steps, we find that steps can become very small when a plateau is reached. 
This can drastically slow down the learning process, and in the worst case cause algorithms to terminate prematurely. 
However, we can use some additional information to refine the gradient. 
Figure \ref{fig:nat-grad-adv} shows an example by Peters et al. \cite{Peters_IICHR_2003} that gives a visual intuition about the difference between 'vanilla' and natural policy gradients. 

Using the Fisher information matrix $F_\theta$, and the gradient estimate we discussed in section \ref{sec:pge} gives us the definition
\begin{equation}
  \widetilde{\nabla}_\theta J(\theta) := F^{-1}_\theta \nabla_\theta J(\theta)
  \label{eqn:nat-grad}
\end{equation}
of the natural gradient. 
The Fisher information matrix represents the certainty we have on our estimate of the gradient and is defined as the covariance of the log likelihood function of a trajectory $\tau_{\pi}^T$, which as Peters et al. \cite{4863} show can be written as 
\begin{equation}
  F_\theta = \int_\mathbb{S} d^\pi(s) \int_\mathbb{A} \pi(a|s) \nabla_\theta \log{\pi(a|s)} \nabla_\theta \log{\pi(a|s)}^T dads.
  \label{eqn:F}
\end{equation}
Using a value function estimator and calculating the natural gradient, we get the natural policy gradient algorithm (NPG) \cite{NIPS2017_7233}.
But, if we recall the definition \eqref{eqn:like-grad} of likelihood-ratio gradients, and the compatible function approximator from \eqref{eqn:compatible-approximator}, we get
\begin{equation}
  \nabla_\theta J(\theta) = F_\theta w.
  \label{eqn:J-equals-F}
\end{equation}

From \eqref{eqn:nat-grad}, and \eqref{eqn:J-equals-F}, it follows that
\begin{equation}
  \widetilde{\nabla}_\theta J(\theta) = F^{-1}_\theta \nabla_\theta J(\theta) = F_\theta^{-1} F_\theta w = w.
\end{equation}
Thus, this approach does not require an actual estimate of the Fisher information matrix, but only an estimate of $w$, with the update step according to $\theta_{k+1} = \theta_k + \alpha_k w$.

Peters et al. \cite{4863} present this idea and suggest LSTD-Q($\lambda$), a version of least-squares temporal difference learning  \cite{Boyan:1999:LTD:645528.657618}, as well as episodic natural actor-critic (eNAC).

\section{Conclusion}
\label{sec:outro}

In this paper, we have introduced policy gradient methods as a class of reinforcement learning algorithms. 
We show why policy gradient methods are effective in these environments, and we give some intuitions for the concept. 
Further, we show the core elements of policy gradient methods, discuss some intricacies the estimation of the policy gradient brings, and follow the research development in the attempts of improving the efficiency and stability of policy gradients.
We show that we can reuse value-estimation approaches in actor-critic settings to improve gradient estimate through better policy evaluation.
This leads to the introduction of the natural gradient as a way to iterate through policy space instead of parameter space, which improves sample efficiency, especially when the gradient in parameter space is very small.

From the developments in recent research, it is fair to say that policy gradient methods play a major role in reinforcement learning. 

\section*{Acknowledgments}

Many thanks to Samuele Tosatto for his helpful reviews. 
Also, this paper would not exist without the engaging lectures on reinforcement learning by Jan Peters.

\bibliographystyle{IEEEtran}
\bibliography{bibliography}

\end{document}